\documentclass[runningheads]{llncs}
\usepackage[T1]{fontenc}
\usepackage{array}
\usepackage{graphicx,verbatim}

\usepackage{booktabs}
\usepackage{multirow}
\usepackage[dvipsnames]{xcolor}
\usepackage{algorithm}
\usepackage{algpseudocode}
\begin{document}
\title{DermAgent: A Self-Reflective Agentic System for Dermatological Image Analysis with Multi-Tool Reasoning and Traceable Decision-Making}
\titlerunning{DermAgent: A Collaborative Agent for Dermatological Image Analysis}
%
\author{Yize Liu\inst{1,2} \and
Siyuan Yan\inst{1,2}\thanks{Corresponding author: siyuan.yan@monash.edu} \and
Ming Hu\inst{1,2} \and
Lie Ju\inst{3} \and
Xieji Li\inst{1,2} \and
Feilong Tang\inst{1,2} \and
Wei Feng\inst{1,2} \and
Zongyuan Ge\inst{1,2}}
%
\authorrunning{Y. Liu et al.}
\institute{AIM for Health Lab, Faculty of Information Technology, Monash University,
Melbourne, Australia \and
Faculty of Information Technology, Monash University, Melbourne, Australia \and
University College London, Institute of Ophthalmology, London, United Kingdom}
  
\maketitle              
\begin{abstract}
Dermatological diagnosis requires integrating fine-grained visual perception with expert clinical knowledge. Although Multimodal Large Language Models (MLLMs) facilitate interactive medical image analysis, their application in dermatology is hindered by insufficient domain-specific grounding and hallucinations. To address these issues, we propose DermAgent, a collaborative multi-tool agent that orchestrates seven specialized vision and language modules within a Plan--Execute--Reflect framework.  DermAgent delivers stepwise, traceable diagnostic reasoning through three core components. First, it employs complementary visual perception tools for comprehensive morphological description, dermoscopic concept annotation, and disease diagnosis. Second, to overcome the lack of domain prior, a dual-modality retrieval module anchors every prediction in external evidence by cross-referencing 413,210 diagnosed image cases and 3,199 clinical guideline chunks. To further mitigate hallucinations, a deterministic critic module conducts strict post-hoc auditing via confidence, coverage, and conflict gates, automatically detecting inter-source disagreements to trigger targeted self-correction. Extensive experiments on five dermatology benchmarks demonstrate that DermAgent consistently outperforms state-of-the-art MLLMs and medical agent baselines across zero-shot fine-grained disease diagnosis, concept annotation, and clinical captioning tasks, exceeding GPT-4o by 17.6\% in skin disease diagnostic accuracy and 3.15\% in captioning ROUGE-L. Our code is available at https://github.com/YizeezLiu/DermAgent.


\end{abstract}

\section{Introduction}

Accurate dermatological diagnosis is a complex process that demands far more than visual pattern recognition~\cite{liopyrisArtificialIntelligenceDermatology2022}. With a vast and long-tailed taxonomy of skin conditions~\cite{estevaDermatologistlevelClassificationSkin2017}, clinicians must integrate subtle morphological cues with ontological knowledge of diagnostic mimics and rare variants. This workflow is inherently multimodal and iterative: it involves formulating differential diagnoses, verifying hypotheses against dermoscopic findings, and cross-referencing these findings with medical literature. Developing automated systems that can replicate not only the outcome of expert diagnosis but also the evidence-based reasoning underlying it remains an open challenge.

Deep learning has achieved performance comparable to that of dermatologists on static classification tasks~\cite{estevaDermatologistlevelClassificationSkin2017,haenssleManMachineDiagnostic2018}, and recent foundation models~\cite{yanMultimodalVisionFoundation2025,kimMONETTransparentMedicalImage2024} have further advanced visual representation learning. However, these models operate primarily as isolated, task-specific systems. This lack of flexibility restricts their utility as interactive diagnostic aids~\cite{haggenmullerPatientsDermatologistsPreferences2024a}. To enable interactive reasoning over diverse clinical queries, researchers have turned to Multimodal Large Language Models (MLLMs)~\cite{openaiGPT4oSystemCard2024,baiQwen3VLTechnicalReport2025,liLLaVAMedTrainingLarge2023,chenHuatuoGPTVisionInjectingMedical2024}. While these systems integrate visual understanding with natural language generation to facilitate open-ended dialogue, their clinical applicability is hindered by two fundamental limitations. First, their reasoning is grounded solely in parametric knowledge acquired during pre-training, with no mechanism to retrieve or verify information against external clinical references such as diagnostic guidelines or curated case repositories. This absence of evidence-based grounding makes them prone to hallucinations~\cite{liuSurveyHallucinationLarge2024}, a particularly critical risk in dermatology where morphologically similar lesions demand precise differential diagnosis~\cite{hagerEvaluationMitigationLimitations2024,pillaiGenerativeArtificialIntelligence2024}. Second, accurate dermatological diagnosis requires integrating signals across diverse examination dimensions. This process encompasses visual pattern classification to map images to candidate diagnoses, structured dermoscopic concept extraction for detecting standardized features (e.g., pigment networks and border irregularity), and free-form morphological description in natural language. Existing MLLMs lack both the specialized perception models needed to capture these domain-specific features and the reasoning mechanisms to detect and resolve the frequent contradictions.

Agentic AI provides a promising approach to address these challenges by orchestrating specialized external tools~\cite{lyuWSIAgentsCollaborativeMultiAgent,zhaoAgenticSystemRare2025,ferberDevelopmentValidationAutonomous2025}. Motivated by this paradigm, we propose \textbf{DermAgent}, a multi-tool collaborative agent designed for comprehensive and interpretable dermatological image analysis. First, to overcome the limited transparency of end-to-end inference in isolated models, DermAgent employs an LLM controller that operates through a Plan--Execute--Reflect loop. This controller decomposes complex clinical queries into structured sub-tasks and dispatches them to specialized vision and language tools, thereby providing stepwise reasoning where intermediate outputs are individually inspectable. Second, to prevent reliance on parametric memory alone and to anchor each reasoning step in verifiable clinical evidence, we equip the system with a dual-modality knowledge retrieval module. This module comprises Case-Based Image Retrieval (Case RAG) and Guideline-Grounded Text Retrieval (Guideline RAG), which provide complementary knowledge sources that cross-validate the visual findings of the agent. Third, to detect and resolve inter-source conflicts and flag low-confidence outputs that arise across these diverse sources, we design a deterministic Critic module. This module performs post-hoc auditing of the assembled evidence chain and directs the agent to collect targeted external validation before finalizing the diagnosis.

Our contributions are as follows: (1) We introduce DermAgent, a collaborative agent that orchestrates seven specialist vision and language tools, providing stepwise, inspectable reasoning. (2) We design a dual-modality retrieval module combining Case RAG over $413{,}210$ diagnosed cases with Guideline RAG over $3{,}199$ clinical guideline chunks, grounding predictions in traceable external knowledge. (3) We design a deterministic Critic module with confidence, coverage, and conflict gates that trigger targeted self-correction to suppress hallucinations. (4) Extensive experiments on five datasets demonstrate consistent improvements over MLLM and agent baselines across disease diagnosis, concept annotation, and clinical captioning.

\section{DermAgent}

\begin{figure}[t!]
\centering
\includegraphics[width=0.95\textwidth]{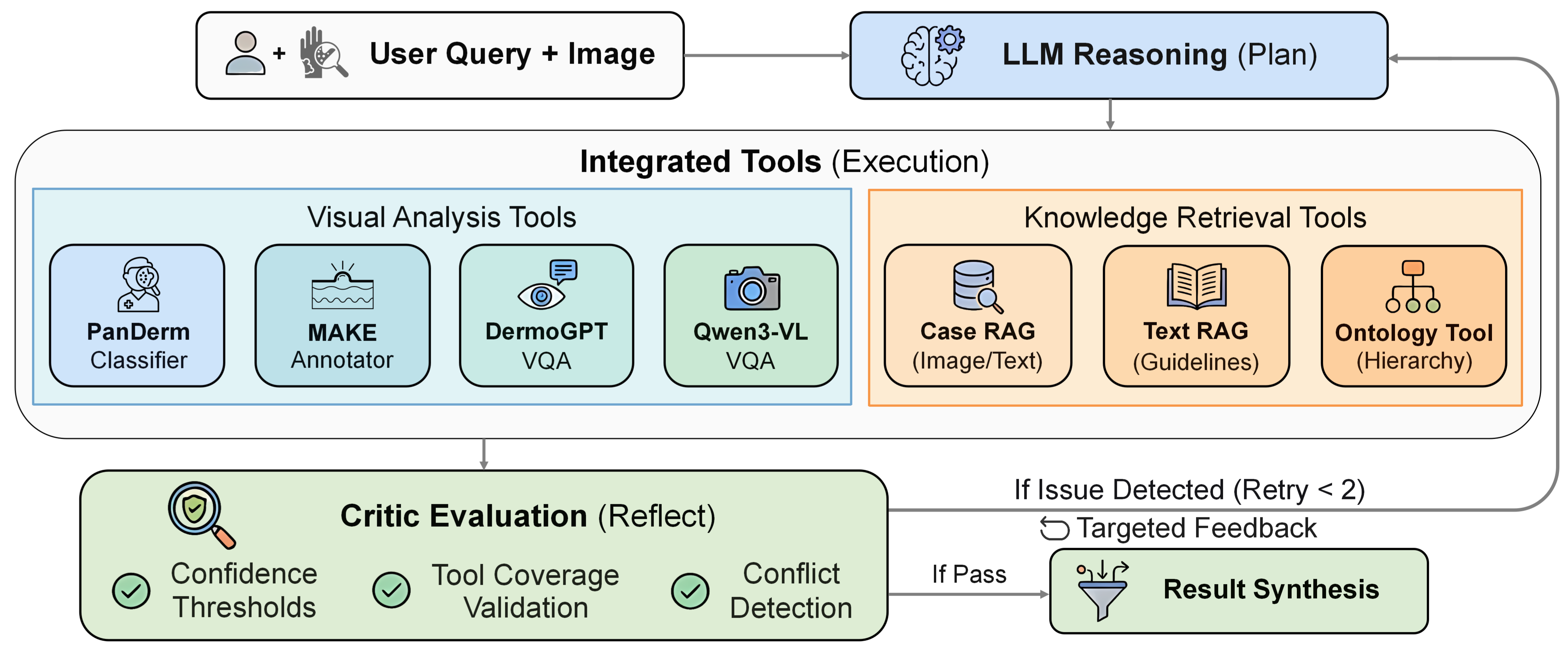}
\caption{Overview of the proposed DermAgent framework. An LLM controller orchestrates specialized visual perception and knowledge retrieval tools via an iterative Plan--Execute--Reflect loop. A deterministic Critic module further audits the accumulated evidence chain to trigger targeted self-correction.} \label{fig:framework}
\end{figure}

We formulate dermatological diagnosis as a multi-step evidence reasoning problem. Given a query tuple $(I, q)$ consisting of a skin image $I$ and a clinical question $q$, the goal is to generate a response $R$ accompanied by a traceable evidence chain $\mathcal{E}$, where each element $e_i \in \mathcal{E}$ represents a verified finding from a specialized tool. As illustrated in Fig.~\ref{fig:framework}, DermAgent operates through a \textit{Plan--Execute--Reflect} loop, dynamically orchestrating seven specialist tools to simulate the clinical workflow of observation, analysis, and reference consultation.

\subsection{Architecture and Workflow}

The agent architecture is modeled as a directed cyclic graph comprising three nodes: \textit{Chatbot}, \textit{Tools}, and \textit{Critic}. The execution flow is formalized in Algorithm~\ref{alg:main}. The system maintains a dynamic evidence chain $\mathcal{E}$ (accumulating verified findings), a feedback memory $\mathcal{G}$ (from the Critic), and a retry counter $k$.

\noindent\textbf{Task Analysis \& Planning.}
Initially, the agent performs task analysis to determine the problem scope $\mathcal{S}$. In the planning phase of iteration $k$, the \textit{Chatbot} node (powered by GPT-4o~\cite{openaiGPT4oSystemCard2024}) synthesizes the current evidence $\mathcal{E}$ and prior critic feedback $\mathcal{G}$ to generate a coherent plan $\mathcal{P}_k = \{(t_i, \theta_i)\}$, consisting of specific tool calls $t_i$ and their parameters $\theta_i$. 

\noindent\textbf{Execute \& Reflect.}
The \textit{Tools} node executes the planned calls in parallel, collecting results $e_i$ and updating the evidence chain: $\mathcal{E} \gets \mathcal{E} \cup \{(t_i, \theta_i, e_i)\}$.
To prevent hallucinations, the \textit{Critic} node acts as a quality controller. It evaluates the updated $\mathcal{E}$ against three logic checks (detailed in Sec.~\ref{sec:critic}). If deficiencies are found (e.g., low confidence or missing coverage) and $k < k_{\max}$, the Critic generates directional feedback $\mathcal{G}$ and routes the flow back to the Chatbot for refinement. Otherwise, the agent proceeds to \textit{Synthesis} to generate the final response $R$.

\begin{algorithm}[t!]
\caption{DermAgent Plan--Execute--Reflect Framework}
\label{alg:main}

\resizebox{\linewidth}{!}{%
    \begin{minipage}{1.15\linewidth} 
    
        \begin{algorithmic}[1]
        \Require $Q$: User query, $I$: Dermatological image, $\mathcal{T}$: Specialist tool set, $k_{\max}$: Maximum retries
        \Ensure $R$: Final response with evidence chain $\mathcal{E}$

        \State $k \gets 0$, \; $\mathcal{E} \gets \emptyset$, \; $\mathcal{G} \gets \emptyset$
        \State $\mathcal{S} \gets \textsc{AnalyzeTask}(Q, I)$ \Comment{Identify task type and requirements}

        \Repeat
        \Statex \quad \textbf{--- Plan ---}
        \State $\mathcal{P}_k \gets \textsc{Plan}(\mathcal{S}, \mathcal{E}, \mathcal{G} \mid \mathcal{T})$ \Comment{LLM selects tools and queries}

        \Statex \quad \textbf{--- Execute ---}
        \For{$(t_i, \theta_i) \in \mathcal{P}_k$} \Comment{$\theta_i$: tool-specific query/parameters}
            \State $e_i \gets \textsc{Execute}(t_i, \theta_i, I)$
            \State $\mathcal{E} \gets \mathcal{E} \cup \{(t_i, \theta_i, e_i)\}$
        \EndFor

        \Statex \quad \textbf{--- Reflect (Critic) ---}
        \State $f_{\mathrm{conf}} \gets \textsc{CheckConfidence}(\mathcal{E})$ \Comment{Low-confidence detection}
        \State $f_{\mathrm{cov}} \gets \textsc{CheckCoverage}(\mathcal{S}, \mathcal{E})$ \Comment{Missing tool detection}
        \State $f_{\mathrm{con}} \gets \textsc{DetectConflicts}(\mathcal{E})$ \Comment{Cross-tool disagreement}

        \If{$(f_{\mathrm{conf}} \lor f_{\mathrm{cov}} \lor f_{\mathrm{con}})$ \textbf{and} $k < k_{\max}$}
            \State $\mathcal{G} \gets \mathcal{G} \cup \textsc{Feedback}(f_{\mathrm{conf}}, f_{\mathrm{cov}}, f_{\mathrm{con}}, \mathcal{E})$ \Comment{Directional critique}
            \State $k \gets k + 1$
        \EndIf

        \Until{$\neg(f_{\mathrm{conf}} \lor f_{\mathrm{cov}} \lor f_{\mathrm{con}})$ \textbf{or} $k \geq k_{\max}$}

        \Statex \textbf{--- Synthesis ---}
        \State $R \gets \textsc{Synthesize}(\mathcal{E}, Q)$ \Comment{Cross-validate and generate evidence-grounded answer}
        \State \Return $R$

        \end{algorithmic}
        
    \end{minipage}%
}

\end{algorithm}

\subsection{Specialist Tool Modules}

DermAgent assembles seven specialist tools organized into two functional groups.
\textit{Visual perception modules} are adopted from established foundation models~\cite{yanMultimodalVisionFoundation2025,yanMAKEMultiAspectKnowledgeEnhanced2025,ruDermoGPTOpenWeights2026,baiQwen3VLTechnicalReport2025}, selected for their complementary representational axes and state-of-the-art performance on dermatological benchmarks.
\textit{Knowledge retrieval mechanisms} ground agent predictions in non-parametric external evidence at inference time, providing verifiable, source-attributed reasoning that cannot be fabricated from model weights alone.

\noindent\textbf{PanDerm Classifier.}
This module performs zero-shot skin disease classification using DermLIP~\cite{yanMultimodalVisionFoundation2025} over a diverse disease taxonomy. Given an image $I$ and a candidate label set $\mathcal{C} = \{c_1, \ldots, c_N\}$, DermLIP computes cosine similarity between the image embedding and the text embedding of each candidate, returning a ranked prediction list with calibrated confidence scores.

\noindent\textbf{MAKE Concept Annotator.}
MAKE~\cite{yanMAKEMultiAspectKnowledgeEnhanced2025} detects dermoscopic concepts (e.g., pigment network, streaks) via zero-shot CLIP-based scoring. For an image $I$ and feature set $\mathcal{F} = \{f_1, \ldots, f_M\}$, it derives visual--semantic alignments to output a structured annotation set $\mathcal{A} \subseteq \mathcal{F}$.

\noindent\textbf{DermoGPT (Dermatology VQA).}
DermoGPT~\cite{ruDermoGPTOpenWeights2026} is a dermatology specialized vision--language model that provides free-form visual question answering for morphological description and diagnostic reasoning. 

\noindent\textbf{Qwen3-VL (General VQA).} We employ Qwen3-VL-8B-Instruct~\cite{baiQwen3VLTechnicalReport2025} specifically for complementary VQA tasks.

\noindent\textbf{Case-Based Image Retrieval (Case RAG).}
The Case RAG module encodes a query image with the DermLIP encoder (shared with the PanDerm classifier) into a 512-dimensional embedding and performs cosine similarity search over a vector database containing 413{,}210 clinically diagnosed cases from Derm1M~\cite{yanDerm1MMillionscaleVisionLanguage2025}. Each retrieved entry carries a disease label, a hierarchical diagnostic category, and a clinical description, providing non-parametric evidence grounded in real diagnostic records. 

\noindent\textbf{Guideline-Grounded Text Retrieval (Guideline RAG).}
The Guideline RAG module queries a curated knowledge base of 3{,}199 document chunks compiled from DermNet~\cite{DermNet} (2{,}751 chunks) and Mayo Clinic~\cite{MayoClinicMedical} (448 chunks) dermatology references. Retrieval follows a four-stage hybrid pipeline. The input query first undergoes domain-specific stop-word filtering to remove generic medical and interrogative terms. The filtered query is then encoded by Qwen3-Embedding-8B~\cite{qwen3embedding} into 4{,}096-dimensional embeddings for semantic vector search, while simultaneously tokenized for full-text keyword matching over indexed fields. The two candidate lists are merged via Reciprocal Rank Fusion (RRF, $k{=}60$) and re-ranked by Qwen3-Reranker-0.6B~\cite{qwen3embedding}, a 0.6-billion-parameter cross-encoder that scores each query--document pair through a binary relevance prompt. The module returns disease names, clinical guideline sections, and source URLs, supplying guideline-grounded evidence for differential diagnosis and terminology verification.

\noindent\textbf{Ontology Tool.} This tool is a hierarchical knowledge graph of skin disease taxonomy with fuzzy name matching. Given a query mode $m \in \{\mathrm{hierarchy}, \allowbreak \mathrm{children}, \allowbreak \mathrm{siblings}, \allowbreak \mathrm{search}\}$ and a disease name $d$, it returns the corresponding set of disease entities or hierarchical relationships.


\subsection{Critic-Driven Reflection}\label{sec:critic}

After each execution round, a deterministic Critic module evaluates the accumulated evidence chain $\mathcal{E}$ against three gate conditions (Algorithm~\ref{alg:main}, lines 11--14). If \emph{any} condition is satisfied and $k < k_{\max}$, the Critic routes control back to the Chatbot node, where the LLM controller re-examines the image and current evidence to autonomously plan a new round of evidence collection. Otherwise, $\mathcal{E}$ is forwarded to final synthesis.

\noindent\textbf{Confidence Check ($f_{\mathrm{conf}}$).} The Critic flags low-confidence outputs based on empirically set thresholds: PanDerm predictions with cosine similarity below 90\%, or RAG retrievals with similarity below 80\%. A flag is raised only when an uninvoked actionable tool remains available, ensuring retries introduce genuinely new evidence rather than repeating prior calls.

\noindent\textbf{Coverage Check ($f_{\mathrm{cov}}$).} The Critic verifies that task-critical specialist tools have been invoked for the given query type. This prevents the LLM controller from short-circuiting to a VQA model alone without engaging task-specific modules (e.g., PanDerm for diagnosis, MAKE for concept annotation), ensuring each final answer is grounded in evidence from multiple complementary sources.

\noindent\textbf{Conflict Check ($f_{\mathrm{con}}$).} The Critic compares top-ranked predictions across visual specialist tools (PanDerm and Case RAG) and flags unresolved disagreements. Conflicts already addressed through prior Guideline RAG retrieval are accepted, preventing unproductive retry loops.

\section{Experiments}

\noindent\textbf{Experimental Setup.} DermAgent is evaluated on five established dermatology benchmarks spanning three complementary tasks. For zero-shot classification: HAM10000~\cite{tschandlHAM10000DatasetLarge2018} (7 diseases, 642 images) and SNU~\cite{hanSNUDatasetQuiz2022} (134 fine-grained disease classes, 500 images). For dermoscopic concept annotation: Derm7pt~\cite{kawaharaSevenPointChecklistSkin2019} (7 dermoscopic concepts, 100 images) and SkinCon~\cite{daneshjouSkinConSkinDisease2023} (32 clinical concepts, 100 images). For clinical captioning: SkinCAP~\cite{shenSkinCaReMultimodalDermatology2025} (100 images). We employ stratified sampling across all benchmarks and oversample underrepresented classes to ensure balanced evaluation coverage.

\subsection{Main Results}
\begin{table}[t!]
\caption{Performance comparison on dermatological diagnosis (Accuracy), concept annotation (F1-Macro), and captioning (ROUGE-L) tasks (\%).}\label{tab:main_results}
\centering
\resizebox{\textwidth}{!}{%
\begin{tabular}{llccccc}
\toprule
\multirow{2}{*}{\textbf{Model}} & \multirow{2}{*}{\textbf{Type}} & \multicolumn{2}{c}{\textbf{Diagnosis}} & \multicolumn{2}{c}{\textbf{Concept Annotation}} & \textbf{Captioning} \\
\cmidrule(lr){3-4} \cmidrule(lr){5-6} \cmidrule(lr){7-7}
& & \textbf{HAM10000} & \textbf{SNU} & \textbf{Derm7pt} & \textbf{SkinCon} & \textbf{SkinCAP} \\
\midrule
LLaVA-Med-v1.5~\cite{liLLaVAMedTrainingLarge2023}      & Medical MLLM & 44.24 & 1.20 & 51.70 & 13.10 & 15.32 \\
HuatuoGPT~\cite{chenHuatuoGPTVisionInjectingMedical2024}           & Medical MLLM & 51.40 & 4.00 & 53.43 & 9.49 & 14.32 \\
Hulu-Med-7B~\cite{jiangHuluMedTransparentGeneralist2025}  & Medical MLLM & 52.65 & 0.80 & 51.63 & 9.63 & 11.43 \\
DermoGPT-RL~\cite{ruDermoGPTOpenWeights2026}          & Dermatology MLLM & 50.00 & 9.20 & 56.86 & 20.72 & 15.41 \\
SkinVL-PubMM~\cite{zengMMSkinEnhancingDermatology2025}         & Dermatology MLLM & 45.17 & 3.40 & 53.14 & 13.20 & 14.44 \\

Qwen3-VL-8B~\cite{baiQwen3VLTechnicalReport2025}         & General MLLM & 51.09 & 7.80 & 53.70 & 22.82 & 12.47 \\
GPT-4o~\cite{openaiGPT4oSystemCard2024}               & General MLLM & 48.91 & 15.00 & 54.14 & 29.56 & 16.33 \\
GPT-5.2~\cite{IntroducingGPT522026}              & General MLLM & 35.98 & 14.80 & 53.86 & 26.62 & 12.35 \\
\midrule
MDAgents~\cite{kimMDAgentsAdaptiveCollaboration2024}             & Medical Agent & 16.82 & 11.40 & 36.14 & 23.93 & 11.99 \\
MedAgent-Pro~\cite{wangMedAgentProEvidencebasedMultimodal2025}         & Medical Agent & 57.63 & 11.60 & 64.82 & 18.34 & 11.48 \\
\textbf{DermAgent (Ours)} & \textbf{Medical Agent} & \textbf{61.83} & \textbf{32.60} & \textbf{65.06} & \textbf{32.95} & \textbf{19.48} \\
\bottomrule
\end{tabular}%
}
\end{table}

Table~\ref{tab:main_results} presents performance comparisons across all benchmarks. In zero-shot classification, DermAgent achieves 61.83\% accuracy on HAM10000, surpassing MedAgent-Pro (57.63\%) by +4.20\% and the strongest MLLM baseline (Hulu-Med-7B, 52.65\%) by +9.18\%. On the fine-grained SNU dataset (134 classes), DermAgent attains 32.60\%, achieving more than twice the accuracy of the next best result (GPT-4o, 15.00\%). For concept annotation, DermAgent achieves the highest F1-Macro on both Derm7pt (65.06\%, versus 56.86\% for DermoGPT-RL, +8.20\%) and SkinCon (32.95\%, versus 29.56\% for GPT-4o, +3.39\%), confirming that cross-validation between structured annotations and visual descriptions effectively mitigates false positives. In clinical captioning, DermAgent obtains a ROUGE-L of 19.48\% on SkinCAP, outperforming GPT-4o (16.33\%) by +3.15\%.

To quantify the contribution of individual components, we conduct a leave-one-out (LOO) ablation study on SkinCAP (Table~\ref{tab:ablation}).
\begin{table}[t!]
\caption{Ablation study on SkinCAP clinical captioning (\%).}\label{tab:ablation}
\centering
\footnotesize
\begin{tabular}{lcc}
\toprule
\textbf{Configuration} & \textbf{ROUGE-L} & \textbf{$\Delta$} \\
\midrule
\textbf{Full Agent (w/ Critic)} & \textbf{19.48} & \textbf{+2.21} \\
\midrule
Full Agent (w/o Critic) & 17.27 & --- \\
\quad w/o Case RAG          & 15.80 & $-$1.47 \\
\quad w/o Guideline RAG     & 16.28 & $-$0.99 \\
\quad w/o DermoGPT          & 16.72 & $-$0.55 \\
\quad w/o PanDerm           & 16.76 & $-$0.51 \\
\quad w/o MAKE              & 16.79 & $-$0.48 \\
\quad w/o Ontology          & 17.12 & $-$0.15 \\
\bottomrule
\end{tabular}
\end{table}

\noindent\textbf{Critic mechanism.} The Critic provides the largest improvement (+2.21\%), confirming that automatic consistency checking and targeted retries yield more coherent and evidence-grounded captions.

\noindent\textbf{Retrieval modules.} Among individual tools, Case RAG ($\Delta = -1.47\%$) and Guideline RAG ($\Delta = -0.99\%$) are the two most impactful components. Case RAG grounds the clinical impression in real diagnostic records, while Guideline RAG supplies precise medical terminology that improves lexical alignment with reference descriptions.

\noindent\textbf{Visual perception and ontology tools.} Removing DermoGPT ($\Delta = -0.55\%$) or PanDerm ($\Delta = -0.51\%$) moderately degrades performance, indicating their complementary roles in providing morphological descriptions and diagnostic hypotheses. MAKE ($\Delta = -0.48\%$) contributes structured dermoscopic features, while the ontology tool has minimal impact on captioning.

\subsection{Qualitative Analysis}
\begin{figure}[htbp]
\centering
\small
\renewcommand{\arraystretch}{1.15}
\begin{tabular}{@{} >{\bfseries}p{2.4cm} p{9cm} @{}}
\toprule
\raisebox{-0.85\height}{\includegraphics[width=0.19\textwidth]{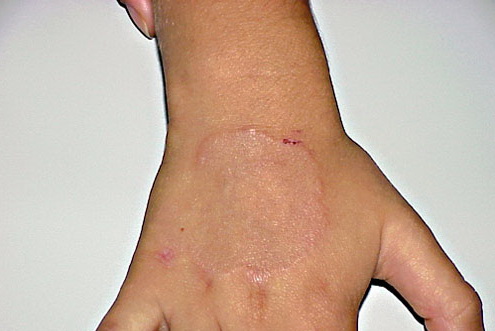}} & \textnormal{\textbf{Ground Truth:}} \textcolor{ForestGreen}{Granuloma annulare} is a relatively rare skin lesion characterized by \textcolor{ForestGreen}{localized red patches with raised ring-shaped edges} and sometimes crusting. This condition may be caused by allergic reactions, infections, or other inflammatory processes\,[\ldots] \\
\midrule
GPT-4o (Baseline) & The skin lesion appears to be a circular patch located on the hand\,[\ldots]\ The borders are distinct but not sharply demarcated\,[\ldots]\ These features are often associated with conditions like \textcolor{BrickRed}{tinea corporis (ringworm) or other types of dermatitis}. \\
\midrule
DermAgent (Ours) & {\scriptsize
\textbf{Round\,1:} DermoGPT $\!\to\!$ \textcolor{ForestGreen}{\textit{erythematous, raised borders, dorsal hand}} $\!\to\!$ \textcolor{BrickRed}{Eczema};\; PanDerm $\!\to\!$ \textcolor{BrickRed}{Eczema\,(100\%)};\; Case\,RAG $\!\to\!$ \textcolor{ForestGreen}{Granuloma Annulare}\,($\!\times\!$4, sim$\!>\!$0.73);\; Guideline\,RAG $\!\to\!$ eczema features\newline
\textit{\textbf{Critic:} $f_{\mathrm{con}}$ conflict\,$\to$\,image reinject\,$\to$\,LLM observes \textcolor{ForestGreen}{round shape, central clearing, raised borders}\,$\to$\,conflicts with Eczema}\newline
\textbf{Round\,2:} PanDerm\,(refined) $\!\to\!$ \textcolor{ForestGreen}{Granuloma Annulare\,(100\%)};\; Guideline\,RAG $\!\to\!$ \textcolor{ForestGreen}{GA clinical features \& guidelines}}\par\smallskip
The image shows a \textcolor{ForestGreen}{round, well-defined lesion on the dorsum of the hand} with \textcolor{ForestGreen}{erythematous, slightly raised borders and central clearing}, consistent with \textcolor{ForestGreen}{granuloma annulare}. Granuloma annulare is a benign inflammatory skin condition\,[\ldots] \\
\bottomrule
\end{tabular}
\caption{Qualitative comparison on a representative case in the captioning task. \textcolor{ForestGreen}{Green} highlights correct descriptions; \textcolor{BrickRed}{red} highlights hallucinated diagnoses.}
\label{fig:qualitative_comparison}
\end{figure}

Figure~\ref{fig:qualitative_comparison} presents a representative case of granuloma annulare. While GPT-4o misidentifies the lesion as tinea corporis, DermAgent overcomes this bias toward common classes through critic-driven self-correction. Initially, DermoGPT accurately describes the morphological features of the lesion but hallucinates a diagnosis of eczema. PanDerm similarly predicts eczema. However, the Case RAG module retrieves four highly similar cases of granuloma annulare. The $f_{\mathrm{con}}$ gate detects this discrepancy and triggers a retry mechanism with image reinjection. Upon re-evaluating the visual evidence, the LLM controller recognizes that the observed raised borders and central clearing conflict with the diagnostic criteria for eczema. Subsequently, a refined query to PanDerm confirms the diagnosis of granuloma annulare, while the Guideline RAG module retrieves the appropriate clinical guidelines. This process provides the precise terminology required to synthesize an accurate and evidence-grounded caption.

\section{Conclusion}

We introduced DermAgent, a collaborative multi-tool agent for comprehensive dermatological image analysis. It orchestrates specialized visual perception modules and complementary Case RAG and Guideline RAG within a Plan--Execute--Reflect framework, bridging data-driven pattern recognition and knowledge-grounded diagnosis. Evaluations across five benchmarks show that DermAgent outperforms state-of-the-art MLLMs and agents in diagnosis, concept annotation, and captioning, with ablation studies confirming the critic-driven self-correction mechanism and retrieval modules as the most critical components.

    

%

%
%
%
\bibliographystyle{splncs04}
\bibliography{mybibliography}

@misc{jiangHuluMedTransparentGeneralist2025,
      title={Hulu-Med: A Transparent Generalist Model towards Holistic Medical Vision-Language Understanding}, 
      author={Songtao Jiang and Yuan Wang and Sibo Song and Tianxiang Hu and Chenyi Zhou and Bin Pu and Yan Zhang and Zhibo Yang and Yang Feng and Joey Tianyi Zhou and Jin Hao and Zijian Chen and Ruijia Wu and Tao Tang and Junhui Lv and Hongxia Xu and Hongwei Wang and Jun Xiao and Bin Feng and Fudong Zhu and Kenli Li and Weidi Xie and Jimeng Sun and Jian Wu and Zuozhu Liu},
      year={2025},
      eprint={2510.08668},
      archivePrefix={arXiv},
      primaryClass={cs.CV},
      url={https://arxiv.org/abs/2510.08668}
}

@misc{baiQwen3VLTechnicalReport2025,
  title = {Qwen3-{{VL Technical Report}}},
  author = {Bai, Shuai and Cai, Yuxuan and Chen, Ruizhe and Chen, Keqin and Chen, Xionghui and Cheng, Zesen and Deng, Lianghao and Ding, Wei and Gao, Chang and Ge, Chunjiang and Ge, Wenbin and Guo, Zhifang and Huang, Qidong and Huang, Jie and Huang, Fei and Hui, Binyuan and Jiang, Shutong and Li, Zhaohai and Li, Mingsheng and Li, Mei and Li, Kaixin and Lin, Zicheng and Lin, Junyang and Liu, Xuejing and Liu, Jiawei and Liu, Chenglong and Liu, Yang and Liu, Dayiheng and Liu, Shixuan and Lu, Dunjie and Luo, Ruilin and Lv, Chenxu and Men, Rui and Meng, Lingchen and Ren, Xuancheng and Ren, Xingzhang and Song, Sibo and Sun, Yuchong and Tang, Jun and Tu, Jianhong and Wan, Jianqiang and Wang, Peng and Wang, Pengfei and Wang, Qiuyue and Wang, Yuxuan and Xie, Tianbao and Xu, Yiheng and Xu, Haiyang and Xu, Jin and Yang, Zhibo and Yang, Mingkun and Yang, Jianxin and Yang, An and Yu, Bowen and Zhang, Fei and Zhang, Hang and Zhang, Xi and Zheng, Bo and Zhong, Humen and Zhou, Jingren and Zhou, Fan and Zhou, Jing and Zhu, Yuanzhi and Zhu, Ke},
  year = 2025,
  month = nov,
  number = {arXiv:2511.21631},
  eprint = {2511.21631},
  primaryclass = {cs},
  publisher = {arXiv},
  doi = {10.48550/arXiv.2511.21631},
  urldate = {2026-02-12},
  archiveprefix = {arXiv}
}

@article{hanSNUDatasetQuiz2022,
author = "Seung Seog Han",
title = "{SNU dataset + Quiz}",
year = "2019",
month = "3",
url ={},
doi = {10.6084/m9.figshare.6454973.v12}
}

@misc{chenHuatuoGPTVisionInjectingMedical2024,
  title = {{{HuatuoGPT-Vision}}, {{Towards Injecting Medical Visual Knowledge}} into {{Multimodal LLMs}} at {{Scale}}},
  author = {Chen, Junying and Gui, Chi and Ouyang, Ruyi and Gao, Anningzhe and Chen, Shunian and Chen, Guiming Hardy and Wang, Xidong and Zhang, Ruifei and Cai, Zhenyang and Ji, Ke and Yu, Guangjun and Wan, Xiang and Wang, Benyou},
  year = 2024,
  month = sep,
  number = {arXiv:2406.19280},
  eprint = {2406.19280},
  primaryclass = {cs},
  publisher = {arXiv},
  doi = {10.48550/arXiv.2406.19280},
  urldate = {2026-02-12},
  archiveprefix = {arXiv}
}

@misc{DermNet,
  title = {{{DermNet}}},
  urldate = {2026-02-13},
  howpublished = {https://dermnetnz.org/}
}

@misc{MayoClinicMedical,
  title = {Mayo {{Clinic}} - {{Medical Diseases}} \& {{Conditions}}},
  journal = {Mayo Clinic},
  urldate = {2026-02-13},
  howpublished = {https://www.mayoclinic.org/diseases-conditions},
  langid = {english}
}

@misc{daneshjouSkinConSkinDisease2023,
  title = {{{SkinCon}}: {{A}} Skin Disease Dataset Densely Annotated by Domain Experts for Fine-Grained Model Debugging and Analysis},
  shorttitle = {{{SkinCon}}},
  author = {Daneshjou, Roxana and Yuksekgonul, Mert and Cai, Zhuo Ran and Novoa, Roberto and Zou, James},
  year = 2023,
  month = feb,
  number = {arXiv:2302.00785},
  eprint = {2302.00785},
  primaryclass = {cs},
  publisher = {arXiv},
  doi = {10.48550/arXiv.2302.00785},
  urldate = {2026-02-12},
  archiveprefix = {arXiv}
}

@article{estevaDermatologistlevelClassificationSkin2017,
  title = {Dermatologist-Level Classification of Skin Cancer with Deep Neural Networks},
  author = {Esteva, Andre and Kuprel, Brett and Novoa, Roberto A. and Ko, Justin and Swetter, Susan M. and Blau, Helen M. and Thrun, Sebastian},
  year = 2017,
  month = feb,
  journal = {Nature},
  volume = {542},
  number = {7639},
  pages = {115--118},
  publisher = {Nature Publishing Group},
  issn = {1476-4687},
  doi = {10.1038/nature21056},
  urldate = {2026-02-12},
  copyright = {2017 Macmillan Publishers Limited, part of Springer Nature. All rights reserved.},
  langid = {english}
}

@article{haenssleManMachineDiagnostic2018,
  title = {Man against Machine: Diagnostic Performance of a Deep Learning Convolutional Neural Network for Dermoscopic Melanoma Recognition in Comparison to 58 Dermatologists},
  shorttitle = {Man against Machine},
  author = {Haenssle, H. A. and Fink, C. and Schneiderbauer, R. and Toberer, F. and Buhl, T. and Blum, A. and Kalloo, A. and Hassen, A. Ben Hadj and Thomas, L. and Enk, A. and Uhlmann, L. and Alt, Christina and Arenbergerova, Monika and Bakos, Renato and Baltzer, Anne and Bertlich, Ines and Blum, Andreas and {Bokor-Billmann}, Therezia and Bowling, Jonathan and Braghiroli, Naira and Braun, Ralph and {Buder-Bakhaya}, Kristina and Buhl, Timo and Cabo, Horacio and Cabrijan, Leo and Cevic, Naciye and Classen, Anna and Deltgen, David and Fink, Christine and Georgieva, Ivelina and {Hakim-Meibodi}, Lara-Elena and Hanner, Susanne and Hartmann, Franziska and Hartmann, Julia and Haus, Georg and Hoxha, Elti and Karls, Raimonds and Koga, Hiroshi and Kreusch, J{\"u}rgen and Lallas, Aimilios and Majenka, Pawel and Marghoob, Ash and Massone, Cesare and Mekokishvili, Lali and Mestel, Dominik and Meyer, Volker and Neuberger, Anna and Nielsen, Kari and Oliviero, Margaret and Pampena, Riccardo and Paoli, John and Pawlik, Erika and Rao, Barbar and Rendon, Adriana and Russo, Teresa and Sadek, Ahmed and Samhaber, Kinga and Schneiderbauer, Roland and Schweizer, Anissa and Toberer, Ferdinand and Trennheuser, Lukas and Vlahova, Lyobomira and Wald, Alexander and Winkler, Julia and W{\"o}lbing, Priscila and Zalaudek, Iris},
  year = 2018,
  month = aug,
  journal = {Annals of Oncology},
  series = {Immune-Related Pathologic Response Criteria},
  volume = {29},
  number = {8},
  pages = {1836--1842},
  issn = {0923-7534},
  doi = {10.1093/annonc/mdy166},
  urldate = {2026-02-12}
}

@article{hagerEvaluationMitigationLimitations2024,
  title = {Evaluation and Mitigation of the Limitations of Large Language Models in Clinical Decision-Making},
  author = {Hager, Paul and Jungmann, Friederike and Holland, Robbie and Bhagat, Kunal and Hubrecht, Inga and Knauer, Manuel and Vielhauer, Jakob and Makowski, Marcus and Braren, Rickmer and Kaissis, Georgios and Rueckert, Daniel},
  year = 2024,
  month = sep,
  journal = {Nature Medicine},
  volume = {30},
  number = {9},
  pages = {2613--2622},
  publisher = {Nature Publishing Group},
  issn = {1546-170X},
  doi = {10.1038/s41591-024-03097-1},
  urldate = {2026-02-12},
  copyright = {2024 The Author(s)},
  langid = {english}
}

@article{kawaharaSevenPointChecklistSkin2019,
  title = {Seven-{{Point Checklist}} and {{Skin Lesion Classification Using Multitask Multimodal Neural Nets}}},
  author = {Kawahara, Jeremy and Daneshvar, Sara and Argenziano, Giuseppe and Hamarneh, Ghassan},
  year = 2019,
  month = mar,
  journal = {IEEE Journal of Biomedical and Health Informatics},
  volume = {23},
  number = {2},
  pages = {538--546},
  issn = {2168-2208},
  doi = {10.1109/JBHI.2018.2824327},
  urldate = {2026-02-12}
}

@misc{kimMDAgentsAdaptiveCollaboration2024,
  title = {{{MDAgents}}: {{An Adaptive Collaboration}} of {{LLMs}} for {{Medical Decision-Making}}},
  shorttitle = {{{MDAgents}}},
  author = {Kim, Yubin and Park, Chanwoo and Jeong, Hyewon and Chan, Yik Siu and Xu, Xuhai and McDuff, Daniel and Lee, Hyeonhoon and Ghassemi, Marzyeh and Breazeal, Cynthia and Park, Hae Won},
  year = 2024,
  month = oct,
  number = {arXiv:2404.15155},
  eprint = {2404.15155},
  primaryclass = {cs},
  publisher = {arXiv},
  doi = {10.48550/arXiv.2404.15155},
  urldate = {2026-02-12},
  archiveprefix = {arXiv}
}

@article{kimMONETTransparentMedicalImage2024,
  title = {Transparent Medical Image {{AI}} via an Image--Text Foundation Model Grounded in Medical Literature},
  author = {Kim, Chanwoo and Gadgil, Soham U. and DeGrave, Alex J. and Omiye, Jesutofunmi A. and Cai, Zhuo Ran and Daneshjou, Roxana and Lee, Su-In},
  year = 2024,
  month = apr,
  journal = {Nature Medicine},
  volume = {30},
  number = {4},
  pages = {1154--1165},
  publisher = {Nature Publishing Group},
  issn = {1546-170X},
  doi = {10.1038/s41591-024-02887-x},
  urldate = {2026-02-12},
  copyright = {2024 The Author(s), under exclusive licence to Springer Nature America, Inc.},
  langid = {english}
}

@misc{liLLaVAMedTrainingLarge2023,
  title = {{{LLaVA-Med}}: {{Training}} a {{Large Language-and-Vision Assistant}} for {{Biomedicine}} in {{One Day}}},
  shorttitle = {{{LLaVA-Med}}},
  author = {Li, Chunyuan and Wong, Cliff and Zhang, Sheng and Usuyama, Naoto and Liu, Haotian and Yang, Jianwei and Naumann, Tristan and Poon, Hoifung and Gao, Jianfeng},
  year = 2023,
  month = jun,
  number = {arXiv:2306.00890},
  eprint = {2306.00890},
  primaryclass = {cs},
  publisher = {arXiv},
  doi = {10.48550/arXiv.2306.00890},
  urldate = {2026-02-12},
  archiveprefix = {arXiv}
}

@misc{liuSurveyHallucinationLarge2024,
  title = {A {{Survey}} on {{Hallucination}} in {{Large Vision-Language Models}}},
  author = {Liu, Hanchao and Xue, Wenyuan and Chen, Yifei and Chen, Dapeng and Zhao, Xiutian and Wang, Ke and Hou, Liping and Li, Rongjun and Peng, Wei},
  year = 2024,
  month = may,
  number = {arXiv:2402.00253},
  eprint = {2402.00253},
  primaryclass = {cs},
  publisher = {arXiv},
  doi = {10.48550/arXiv.2402.00253},
  urldate = {2026-02-12},
  archiveprefix = {arXiv}
}

@article{lyuWSIAgentsCollaborativeMultiAgent,
  title = {{{WSI-Agents}}: {{A Collaborative Multi-Agent System}} for {{Multi-Modal Whole Slide Image Analysis}}},
  author = {Lyu, Xinheng and Liang, Yuci and Chen, Wenting and Ding, Meidan and Yang, Jiaqi and Huang, Guolin and Zhang, Daokun and He, Xiangjian and Shen, Linlin},
  langid = {english}
}

@misc{openaiGPT4oSystemCard2024,
  title = {{{GPT-4o System Card}}},
  author = {OpenAI and Hurst, Aaron and Lerer, Adam and Goucher, Adam P. and Perelman, Adam and Ramesh, Aditya and Clark, Aidan and Ostrow, A. J. and Welihinda, Akila and Hayes, Alan and Radford, Alec and M{\k a}dry, Aleksander and {Baker-Whitcomb}, Alex and Beutel, Alex and Borzunov, Alex and Carney, Alex and Chow, Alex and Kirillov, Alex and Nichol, Alex and Paino, Alex and Renzin, Alex and Passos, Alex Tachard and Kirillov, Alexander and Christakis, Alexi and Conneau, Alexis and Kamali, Ali and Jabri, Allan and Moyer, Allison and Tam, Allison and Crookes, Amadou and Tootoochian, Amin and Tootoonchian, Amin and Kumar, Ananya and Vallone, Andrea and Karpathy, Andrej and Braunstein, Andrew and Cann, Andrew and Codispoti, Andrew and Galu, Andrew and Kondrich, Andrew and Tulloch, Andrew and Mishchenko, Andrey and Baek, Angela and Jiang, Angela and Pelisse, Antoine and Woodford, Antonia and Gosalia, Anuj and Dhar, Arka and Pantuliano, Ashley and Nayak, Avi and Oliver, Avital and Zoph, Barret and Ghorbani, Behrooz and Leimberger, Ben and Rossen, Ben and Sokolowsky, Ben and Wang, Ben and Zweig, Benjamin and Hoover, Beth and Samic, Blake and McGrew, Bob and Spero, Bobby and Giertler, Bogo and Cheng, Bowen and Lightcap, Brad and Walkin, Brandon and Quinn, Brendan and Guarraci, Brian and Hsu, Brian and Kellogg, Bright and Eastman, Brydon and Lugaresi, Camillo and Wainwright, Carroll and Bassin, Cary and Hudson, Cary and Chu, Casey and Nelson, Chad and Li, Chak and Shern, Chan Jun and Conger, Channing and Barette, Charlotte and Voss, Chelsea and Ding, Chen and Lu, Cheng and Zhang, Chong and Beaumont, Chris and Hallacy, Chris and Koch, Chris and Gibson, Christian and Kim, Christina and Choi, Christine and McLeavey, Christine and Hesse, Christopher and Fischer, Claudia and Winter, Clemens and Czarnecki, Coley and Jarvis, Colin and Wei, Colin and Koumouzelis, Constantin and Sherburn, Dane and Kappler, Daniel and Levin, Daniel and Levy, Daniel and Carr, David and Farhi, David and Mely, David and Robinson, David and Sasaki, David and Jin, Denny and Valladares, Dev and Tsipras, Dimitris and Li, Doug and Nguyen, Duc Phong and Findlay, Duncan and Oiwoh, Edede and Wong, Edmund and Asdar, Ehsan and Proehl, Elizabeth and Yang, Elizabeth and Antonow, Eric and Kramer, Eric and Peterson, Eric and Sigler, Eric and Wallace, Eric and Brevdo, Eugene and Mays, Evan and Khorasani, Farzad and Such, Felipe Petroski and Raso, Filippo and Zhang, Francis and von Lohmann, Fred and Sulit, Freddie and Goh, Gabriel and Oden, Gene and Salmon, Geoff and Starace, Giulio and Brockman, Greg and Salman, Hadi and Bao, Haiming and Hu, Haitang and Wong, Hannah and Wang, Haoyu and Schmidt, Heather and Whitney, Heather and Jun, Heewoo and Kirchner, Hendrik and Pinto, Henrique Ponde de Oliveira and Ren, Hongyu and Chang, Huiwen and Chung, Hyung Won and Kivlichan, Ian and O'Connell, Ian and O'Connell, Ian and Osband, Ian and Silber, Ian and Sohl, Ian and Okuyucu, Ibrahim and Lan, Ikai and Kostrikov, Ilya and Sutskever, Ilya and Kanitscheider, Ingmar and Gulrajani, Ishaan and Coxon, Jacob and Menick, Jacob and Pachocki, Jakub and Aung, James and Betker, James and Crooks, James and Lennon, James and Kiros, Jamie and Leike, Jan and Park, Jane and Kwon, Jason and Phang, Jason and Teplitz, Jason and Wei, Jason and Wolfe, Jason and Chen, Jay and Harris, Jeff and Varavva, Jenia and Lee, Jessica Gan and Shieh, Jessica and Lin, Ji and Yu, Jiahui and Weng, Jiayi and Tang, Jie and Yu, Jieqi and Jang, Joanne and Candela, Joaquin Quinonero and Beutler, Joe and Landers, Joe and Parish, Joel and Heidecke, Johannes and Schulman, John and Lachman, Jonathan and McKay, Jonathan and Uesato, Jonathan and Ward, Jonathan and Kim, Jong Wook and Huizinga, Joost and Sitkin, Jordan and Kraaijeveld, Jos and Gross, Josh and Kaplan, Josh and Snyder, Josh and Achiam, Joshua and Jiao, Joy and Lee, Joyce and Zhuang, Juntang and Harriman, Justyn and Fricke, Kai and Hayashi, Kai and Singhal, Karan and Shi, Katy and Karthik, Kavin and Wood, Kayla and Rimbach, Kendra and Hsu, Kenny and Nguyen, Kenny and {Gu-Lemberg}, Keren and Button, Kevin and Liu, Kevin and Howe, Kiel and Muthukumar, Krithika and Luther, Kyle and Ahmad, Lama and Kai, Larry and Itow, Lauren and Workman, Lauren and Pathak, Leher and Chen, Leo and Jing, Li and Guy, Lia and Fedus, Liam and Zhou, Liang and Mamitsuka, Lien and Weng, Lilian and McCallum, Lindsay and Held, Lindsey and Ouyang, Long and Feuvrier, Louis and Zhang, Lu and Kondraciuk, Lukas and Kaiser, Lukasz and Hewitt, Luke and Metz, Luke and Doshi, Lyric and Aflak, Mada and Simens, Maddie and Boyd, Madelaine and Thompson, Madeleine and Dukhan, Marat and Chen, Mark and Gray, Mark and Hudnall, Mark and Zhang, Marvin and Aljubeh, Marwan and Litwin, Mateusz and Zeng, Matthew and Johnson, Max and Shetty, Maya and Gupta, Mayank and Shah, Meghan and Yatbaz, Mehmet and Yang, Meng Jia and Zhong, Mengchao and Glaese, Mia and Chen, Mianna and Janner, Michael and Lampe, Michael and Petrov, Michael and Wu, Michael and Wang, Michele and Fradin, Michelle and Pokrass, Michelle and Castro, Miguel and de Castro, Miguel Oom Temudo and Pavlov, Mikhail and Brundage, Miles and Wang, Miles and Khan, Minal and Murati, Mira and Bavarian, Mo and Lin, Molly and Yesildal, Murat and Soto, Nacho and Gimelshein, Natalia and Cone, Natalie and Staudacher, Natalie and Summers, Natalie and LaFontaine, Natan and Chowdhury, Neil and Ryder, Nick and Stathas, Nick and Turley, Nick and Tezak, Nik and Felix, Niko and Kudige, Nithanth and Keskar, Nitish and Deutsch, Noah and Bundick, Noel and Puckett, Nora and Nachum, Ofir and Okelola, Ola and Boiko, Oleg and Murk, Oleg and Jaffe, Oliver and Watkins, Olivia and Godement, Olivier and {Campbell-Moore}, Owen and Chao, Patrick and McMillan, Paul and Belov, Pavel and Su, Peng and Bak, Peter and Bakkum, Peter and Deng, Peter and Dolan, Peter and Hoeschele, Peter and Welinder, Peter and Tillet, Phil and Pronin, Philip and Tillet, Philippe and Dhariwal, Prafulla and Yuan, Qiming and Dias, Rachel and Lim, Rachel and Arora, Rahul and Troll, Rajan and Lin, Randall and Lopes, Rapha Gontijo and Puri, Raul and Miyara, Reah and Leike, Reimar and Gaubert, Renaud and Zamani, Reza and Wang, Ricky and Donnelly, Rob and Honsby, Rob and Smith, Rocky and Sahai, Rohan and Ramchandani, Rohit and Huet, Romain and Carmichael, Rory and Zellers, Rowan and Chen, Roy and Chen, Ruby and Nigmatullin, Ruslan and Cheu, Ryan and Jain, Saachi and Altman, Sam and Schoenholz, Sam and Toizer, Sam and Miserendino, Samuel and Agarwal, Sandhini and Culver, Sara and Ethersmith, Scott and Gray, Scott and Grove, Sean and Metzger, Sean and Hermani, Shamez and Jain, Shantanu and Zhao, Shengjia and Wu, Sherwin and Jomoto, Shino and Wu, Shirong and Shuaiqi and Xia and Phene, Sonia and Papay, Spencer and Narayanan, Srinivas and Coffey, Steve and Lee, Steve and Hall, Stewart and Balaji, Suchir and Broda, Tal and Stramer, Tal and Xu, Tao and Gogineni, Tarun and Christianson, Taya and Sanders, Ted and Patwardhan, Tejal and Cunninghman, Thomas and Degry, Thomas and Dimson, Thomas and Raoux, Thomas and Shadwell, Thomas and Zheng, Tianhao and Underwood, Todd and Markov, Todor and Sherbakov, Toki and Rubin, Tom and Stasi, Tom and Kaftan, Tomer and Heywood, Tristan and Peterson, Troy and Walters, Tyce and Eloundou, Tyna and Qi, Valerie and Moeller, Veit and Monaco, Vinnie and Kuo, Vishal and Fomenko, Vlad and Chang, Wayne and Zheng, Weiyi and Zhou, Wenda and Manassra, Wesam and Sheu, Will and Zaremba, Wojciech and Patil, Yash and Qian, Yilei and Kim, Yongjik and Cheng, Youlong and Zhang, Yu and He, Yuchen and Zhang, Yuchen and Jin, Yujia and Dai, Yunxing and Malkov, Yury},
  year = 2024,
  month = oct,
  number = {arXiv:2410.21276},
  eprint = {2410.21276},
  primaryclass = {cs},
  publisher = {arXiv},
  doi = {10.48550/arXiv.2410.21276},
  urldate = {2026-02-12},
  archiveprefix = {arXiv}
}

@article{pillaiGenerativeArtificialIntelligence2024,
  title = {Generative Artificial Intelligence in Dermatology: {{Recommendations}} for Future Studies Evaluating the Clinical Knowledge of Models},
  shorttitle = {Generative Artificial Intelligence in Dermatology},
  author = {Pillai, Joshua and Li, Benjamin},
  year = 2024,
  month = jul,
  journal = {Skin Research and Technology},
  volume = {30},
  number = {7},
  pages = {e13854},
  issn = {0909-752X},
  doi = {10.1111/srt.13854},
  urldate = {2026-02-12},
  pmcid = {PMC11245338},
  pmid = {38997219}
}

@misc{ruDermoGPTOpenWeights2026,
  title = {{{DermoGPT}}: {{Open Weights}} and {{Open Data}} for {{Morphology-Grounded Dermatological Reasoning MLLMs}}},
  shorttitle = {{{DermoGPT}}},
  author = {Ru, Jinghan and Yan, Siyuan and Yin, Yuguo and Zou, Yuexian and Ge, Zongyuan},
  year = 2026,
  month = jan,
  number = {arXiv:2601.01868},
  eprint = {2601.01868},
  primaryclass = {cs},
  publisher = {arXiv},
  doi = {10.48550/arXiv.2601.01868},
  urldate = {2026-02-12},
  archiveprefix = {arXiv}
}

@misc{shenSkinCaReMultimodalDermatology2025,
  title = {{{SkinCaRe}}: {{A Multimodal Dermatology Dataset Annotated}} with {{Medical Caption}} and {{Chain-of-Thought Reasoning}}},
  shorttitle = {{{SkinCaRe}}},
  author = {Shen, Yuhao and Sun, Liyuan and Xu, Yan and Liu, Wenbin and Zhang, Shuping and Afvari, Shawn and Han, Zhongyi and Song, Jiaoyan and Ji, Yongzhi and Lu, Tao and He, Xiaonan and Gao, Xin and Zhou, Juexiao},
  year = 2025,
  month = nov,
  number = {arXiv:2405.18004},
  eprint = {2405.18004},
  primaryclass = {cs},
  publisher = {arXiv},
  doi = {10.48550/arXiv.2405.18004},
  urldate = {2026-02-12},
  archiveprefix = {arXiv}
}

@article{tschandlHAM10000DatasetLarge2018,
  title = {The {{HAM10000}} Dataset, a Large Collection of Multi-Source Dermatoscopic Images of Common Pigmented Skin Lesions},
  author = {Tschandl, Philipp and Rosendahl, Cliff and Kittler, Harald},
  year = 2018,
  month = aug,
  journal = {Scientific Data},
  volume = {5},
  number = {1},
  pages = {180161},
  publisher = {Nature Publishing Group},
  issn = {2052-4463},
  doi = {10.1038/sdata.2018.161},
  urldate = {2026-02-12},
  copyright = {2018 The Author(s)},
  langid = {english}
}

@misc{yanDerm1MMillionscaleVisionLanguage2025,
  title = {{{Derm1M}}: {{A Million-scale Vision-Language Dataset Aligned}} with {{Clinical Ontology Knowledge}} for {{Dermatology}}},
  shorttitle = {{{Derm1M}}},
  author = {Yan, Siyuan and Hu, Ming and Jiang, Yiwen and Li, Xieji and Fei, Hao and Tschandl, Philipp and Kittler, Harald and Ge, Zongyuan},
  year = 2025,
  month = apr,
  number = {arXiv:2503.14911},
  eprint = {2503.14911},
  primaryclass = {cs},
  publisher = {arXiv},
  doi = {10.48550/arXiv.2503.14911},
  urldate = {2026-02-12},
  archiveprefix = {arXiv}
}

@misc{yanMAKEMultiAspectKnowledgeEnhanced2025,
  title = {{{MAKE}}: {{Multi-Aspect Knowledge-Enhanced Vision-Language Pretraining}} for {{Zero-shot Dermatological Assessment}}},
  shorttitle = {{{MAKE}}},
  author = {Yan, Siyuan and Li, Xieji and Hu, Ming and Jiang, Yiwen and Yu, Zhen and Ge, Zongyuan},
  year = 2025,
  month = may,
  number = {arXiv:2505.09372},
  eprint = {2505.09372},
  primaryclass = {cs},
  publisher = {arXiv},
  doi = {10.48550/arXiv.2505.09372},
  urldate = {2026-02-12},
  archiveprefix = {arXiv}
}

@article{yanMultimodalVisionFoundation2025,
  title = {A Multimodal Vision Foundation Model for Clinical Dermatology},
  author = {Yan, Siyuan and Yu, Zhen and Primiero, Clare and {Vico-Alonso}, Cristina and Wang, Zhonghua and Yang, Litao and Tschandl, Philipp and Hu, Ming and Ju, Lie and Tan, Gin and Tang, Vincent and Ng, Aik Beng and Powell, David and Bonnington, Paul and See, Simon and Magnaterra, Elisabetta and Ferguson, Peter and Nguyen, Jennifer and Guitera, Pascale and Banuls, Jose and Janda, Monika and Mar, Victoria and Kittler, Harald and Soyer, H. Peter and Ge, Zongyuan},
  year = 2025,
  month = aug,
  journal = {Nature Medicine},
  volume = {31},
  number = {8},
  pages = {2691--2702},
  publisher = {Nature Publishing Group},
  issn = {1546-170X},
  doi = {10.1038/s41591-025-03747-y},
  urldate = {2026-02-12},
  copyright = {2025 The Author(s)},
  langid = {english}
}

@misc{zhaoAgenticSystemRare2025,
  title = {An {{Agentic System}} for {{Rare Disease Diagnosis}} with {{Traceable Reasoning}}},
  author = {Zhao, Weike and Wu, Chaoyi and Fan, Yanjie and Zhang, Xiaoman and Qiu, Pengcheng and Sun, Yuze and Zhou, Xiao and Wang, Yanfeng and Sun, Xin and Zhang, Ya and Yu, Yongguo and Sun, Kun and Xie, Weidi},
  year = 2025,
  month = aug,
  number = {arXiv:2506.20430},
  eprint = {2506.20430},
  primaryclass = {cs},
  publisher = {arXiv},
  doi = {10.48550/arXiv.2506.20430},
  urldate = {2026-01-28},
  archiveprefix = {arXiv},
  langid = {english}
}

@article{ferberDevelopmentValidationAutonomous2025,
  title = {Development and Validation of an Autonomous Artificial Intelligence Agent for Clinical Decision-Making in Oncology},
  author = {Ferber, Dyke and El Nahhas, Omar SM and W{\"o}lflein, Georg and Wiest, Isabella C. and Clusmann, Jan and Le{\ss}mann, Marie-Elisabeth and Foersch, Sebastian and Lammert, Jacqueline and Tschochohei, Maximilian and J{\"a}ger, Dirk},
  year = 2025,
  journal = {Nature cancer},
  pages = {1--13},
  publisher = {Nature Publishing Group US New York},
  urldate = {2026-01-28}
}

@article{haggenmullerPatientsDermatologistsPreferences2024a,
  title = {Patients' and Dermatologists' Preferences in Artificial Intelligence-Driven Skin Cancer Diagnostics: {{A}} Prospective Multicentric Survey Study},
  shorttitle = {Patients' and Dermatologists' Preferences in Artificial Intelligence-Driven Skin Cancer Diagnostics},
  author = {Haggenm{\"u}ller, Sarah and Maron, Roman C. and Hekler, Achim and {Krieghoff-Henning}, Eva and Utikal, Jochen S. and Gaiser, Maria and M{\"u}ller, Verena and Fabian, Sascha and Meier, Friedegund and Hobelsberger, Sarah and Gellrich, Frank F. and Sergon, Mildred and Hauschild, Axel and Weichenthal, Michael and French, Lars E. and Heinzerling, Lucie and Schlager, Justin G. and Ghoreschi, Kamran and Schlaak, Max and Hilke, Franz J. and Poch, Gabriela and Korsing, S{\"o}ren and Berking, Carola and Heppt, Markus V. and Erdmann, Michael and Haferkamp, Sebastian and Drexler, Konstantin and Schadendorf, Dirk and Sondermann, Wiebke and Goebeler, Matthias and Schilling, Bastian and Kather, Jakob N. and Fr{\"o}hling, Stefan and Kaminski, Katharina and Doppler, Astrid and Bucher, Tabea and Brinker, Titus J. and {Collaborators}},
  year = 2024,
  month = aug,
  journal = {Journal of the American Academy of Dermatology},
  volume = {91},
  number = {2},
  pages = {366--370},
  issn = {1097-6787},
  doi = {10.1016/j.jaad.2024.04.033},
  langid = {english},
  pmid = {38670313}
}

@article{liopyrisArtificialIntelligenceDermatology2022,
  title = {Artificial {{Intelligence}} in {{Dermatology}}: {{Challenges}} and {{Perspectives}}},
  shorttitle = {Artificial {{Intelligence}} in {{Dermatology}}},
  author = {Liopyris, Konstantinos and Gregoriou, Stamatios and Dias, Julia and Stratigos, Alexandros J.},
  year = 2022,
  month = oct,
  journal = {Dermatology and Therapy},
  volume = {12},
  number = {12},
  pages = {2637--2651},
  issn = {2193-8210},
  doi = {10.1007/s13555-022-00833-8},
  urldate = {2026-02-12},
  pmcid = {PMC9674813},
  pmid = {36306100}
}

@article{qwen3embedding,
  title={Qwen3 Embedding: Advancing Text Embedding and Reranking Through Foundation Models},
  author={Zhang, Yanzhao and Li, Mingxin and Long, Dingkun and Zhang, Xin and Lin, Huan and Yang, Baosong and Xie, Pengjun and Yang, An and Liu, Dayiheng and Lin, Junyang and Huang, Fei and Zhou, Jingren},
  journal={arXiv preprint arXiv:2506.05176},
  year={2025}
}

@misc{wangMedAgentProEvidencebasedMultimodal2025,
  title = {{{MedAgent-Pro}}: {{Towards Evidence-based Multi-modal Medical Diagnosis}} via {{Reasoning Agentic Workflow}}},
  shorttitle = {{{MedAgent-Pro}}},
  author = {Wang, Ziyue and Wu, Junde and Cai, Linghan and Low, Chang Han and Yang, Xihong and Li, Qiaxuan and Jin, Yueming},
  year = 2025,
  month = jul,
  number = {arXiv:2503.18968},
  eprint = {2503.18968},
  primaryclass = {cs},
  publisher = {arXiv},
  doi = {10.48550/arXiv.2503.18968},
  archiveprefix = {arXiv},
}

@inproceedings{zengMMSkinEnhancingDermatology2025,
  title = {{{MM-Skin}}: {{Enhancing Dermatology Vision-Language Model}} with an {{Image-Text Dataset Derived}} from {{Textbooks}}},
  shorttitle = {{{MM-Skin}}},
  booktitle = {Proceedings of the 33rd {{ACM International Conference}} on {{Multimedia}}},
  author = {Zeng, Wenqi and Sun, Yuqi and Ma, Chenxi and Tan, Weimin and Yan, Bo},
  year = 2025,
  month = oct,
  series = {{{MM}} '25},
  pages = {3769--3778},
  publisher = {Association for Computing Machinery},
  address = {New York, NY, USA},
  doi = {10.1145/3746027.3755187},
}

@misc{IntroducingGPT522026,
  title = {Introducing {{GPT-5}}.2},
  author = {OpenAI},
  year = 2026,
  month = feb,
  urldate = {2026-02-18},
  howpublished = {https://openai.com/index/introducing-gpt-5-2/},
  langid = {american}
}

\end{document}